\documentclass[11pt]{article}

\PassOptionsToPackage{table}{xcolor}
\usepackage{acl}

\usepackage{times}
\usepackage{latexsym}
\usepackage[T1]{fontenc}
\usepackage[utf8]{inputenc}
\usepackage{microtype}
\usepackage{inconsolata}

\usepackage{graphicx}
\usepackage{amsmath}
\usepackage{amssymb}
\usepackage{amsthm}
\usepackage{booktabs}
\usepackage{multirow}
\usepackage{url}

\title{SimCRAFT: Distilling Remote Sensing Agents via Synthetic Trajectories and Contextual Retrieval-Augmented Fine-Tuning}

\author{
  \textbf{Haoran Wang\textsuperscript{1,2}},
  \textbf{Jing Yao\textsuperscript{1}},
  \textbf{Xu Yang\textsuperscript{1,2}},
  \textbf{Zeqing Wang\textsuperscript{1,2}},
  \\
  \textbf{Yang Zhang\textsuperscript{3}},
  \textbf{Pedram Ghamisi\textsuperscript{4}},
  \textbf{Zhengchao Chen\textsuperscript{1}}
  \\
  {\textsuperscript{1}State Key Laboratory of Remote Sensing and Digital Earth,}
  \\
  {Aerospace Information Research Institute, Chinese Academy of Sciences, Beijing 100101, China}
  \\
  {\textsuperscript{2}University of Chinese Academy of Sciences, Beijing 100049, China}
  \\
  {\textsuperscript{3}The Hong Kong University of Science and Technology (Guangzhou), Guangzhou 511453, China}
  \\
  {\textsuperscript{4}Helmholtz-Zentrum Dresden-Rossendorf, Freiberg 09599, Germany}
  \\
  {\texttt{\{wanghaoran23, yangxu252, wangzeqing22\}@mails.ucas.ac.cn, yaojing@aircas.ac.cn}}
  \\
  {\texttt{yzhang971@connect.hkust-gz.edu.cn, p.ghamisi@hzdr.de, chenzc@radi.ac.cn}}
}

\begin{document}
\maketitle
\begin{abstract}
The unprecedented surge in Earth observation data volume and diversity has exposed a critical bottleneck for traditional manual workflows, catalyzing the emergence of Remote Sensing (RS) Agents. However, the practical deployment of these advanced agents is severely hindered by their heavy reliance on large-scale general-purpose LLMs, which lack deep domain expertise and impose prohibitive infrastructure demands. To resolve this, we propose \textbf{SimCRAFT}, a model-agnostic framework that distills sophisticated RS orchestration capabilities into a compact 7B-scale model. Addressing data scarcity, we first pair a multi-agent synthesis engine with a Mock Execution Engine that checks schema correctness, inter-tool dependencies, and sensor/tool compatibility, producing \textbf{SimRS-14k}, a large-scale, constraint-validated workflow planning corpus.
Second, we propose \textbf{Contextual Retrieval-Augmented Fine-Tuning (CRAFT)} that fine-tunes the model to reason analogically by adapting retrieved Standard Operating Procedures to novel queries under a noise-robust objective, generalizing RAFT to multi-step RS workflow planning without mechanical copying.
Extensive experiments demonstrate that SimCRAFT-7B significantly outperforms open-weights LLMs and rivals advanced closed-source models and specialized RS agents, while reproducing across three 7B backbones. This work contributes a competitive open-weights baseline for lightweight RS intelligence, enabling efficient autonomous deployment under resource-constrained or resource-conserving conditions\footnote{Our source code is available at \url{https://github.com/WangHrYii/SimCRAFT}.}.
\end{abstract}

\section{Introduction}
\label{sec:introduction}

\begin{figure}[!t]
    \centering
    \includegraphics[width=\linewidth]{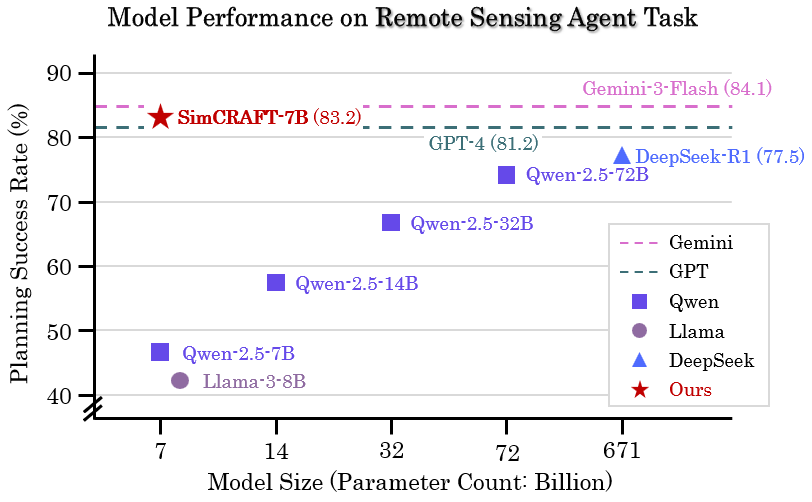}
    \caption{\textbf{Performance vs.\ model size.} SimCRAFT-Qwen2.5-7B (red star) compared with open-weights and closed-source baselines.}
    \label{fig:teaser}
\end{figure}

Earth observation produces petabytes of multi-source imagery each day~\citep{zhang2019remotely,zhang2022progress}. To extract actionable insight from this data, analysts manually compose GDAL, SNAP, and Google Earth Engine scripts that chain a dozen tools such as radiometric calibration, atmospheric correction, and change detection. Yet this process is expert-only, error-prone, and not reusable, which has motivated the emergence of \textbf{Remote Sensing (RS) Agents} that take a natural-language query and decompose it into an executable sequence of tool calls.
The workflow is expert-only, error-prone, and not reusable, which has motivated the emergence of \textbf{Remote Sensing (RS) Agents} that take a natural-language query and decompose it into an executable sequence of tool calls. For example, given the query \emph{``Check illegal excavation near the Terracotta Army site between 2022 and 2026''}, a competent RS agent is expected to plan a 7-step trajectory: RS data search $\rightarrow$ radiometric calibration $\rightarrow$ atmospheric correction $\rightarrow$ temporal change detection $\rightarrow$ vector analysis $\rightarrow$ area statistics. State-of-the-art systems such as EarthAgent~\citep{li2025designing,feng2025earthagent} and CangLing-KnowFlow~\citep{chen2025cangling} achieve this by orchestrating GPT-4-scale closed-source LLMs at runtime. Such a strong dependency tends to result in high compute requirements, strict network connectivity constraints, and increased privacy exposure, precluding their edge deployment on satellites, UAVs, or workstations without high-end GPUs.

Distilling RS Agents into a small-scale open-source backbone becomes a natural target for edge deployment, yet two ready-made paths fail to deliver. \textbf{Path A, inference-time Retrieval-Augmented Generation (RAG)}: a frozen small model fetches procedural templates from an external knowledge base and concatenates them into the prompt. The retrieved unit here can be a Standard Operating Procedure (SOP), a step-by-step template that chains RS tools for a given class of queries. This path yields only marginal gains at, for example, 7B scale, since a frozen small model has limited capacity to robustly recover an executable toolchain from heterogeneous, noisy SOPs. \textbf{Path B, trajectory-only supervised fine-tuning}: train directly on synthetic or expert trajectories. Trajectories are instance-level supervision that ties specific satellites, ROIs, and parameter values to the tool sequence. Small models tend to memorize individual instances rather than abstract the program schema mapping task type to tool composition. This granularity mismatch is especially pronounced in RS, where the parameter space is large, and sensor combinations are diverse. In addition, the RS domain currently lacks a large-scale tool use trajectory corpus that would be needed to drive such supervised fine-tuning (SFT) in the first place.

The common shortcoming of the two paths is that neither teaches the model how to plan tool calls under schema-level SOP guidance. A direction worth exploring is to move SOP retrieval from inference time into training time, so that the model learns each training trajectory alongside its corresponding SOP template and noise perturbations.

We instantiate this direction as \textbf{SimCRAFT}, a model-agnostic two-phase framework. The first phase, \textbf{Multi-Agent Data Synthesis}, addresses the scarcity of large-scale, parameter-precise, and physically consistent training corpora in RS: starting from expert-seeded tasks, it combines dual-mode user simulation, four-agent collaboration, and a Mock Execution Engine with constraint validation over schema, dependencies, and sensor compatibility to produce \textbf{SimRS-14k}, a large-scale, constraint-validated corpus of RS tool-call trajectories. The second phase, \textbf{Contextual Retrieval-Augmented Fine-Tuning (CRAFT)}, addresses the training mechanism: it adapts retrieval-augmented fine-tuning to multi-step RS workflow planning, using two perturbations, Irrelevant Context Injection and Parameter Mutation, that lead the small model to extract transferable procedural structure from noisy SOPs rather than copy them mechanically. The task itself is formalized in Section~\ref{sec:task-formulation} as constrained multi-step tool planning over an action space of 36 RS-specific tools.

Our three-fold contributions are as follows:
\begin{itemize}
    \item \textbf{SimCRAFT framework.} A pure training-time path of data synthesis plus schema-aware supervised fine-tuning that lifts a 7B-scale open-source model to the level of GPT-4-driven RS agents on RS tool use.
    \item \textbf{CRAFT training mechanism.} We adapt retrieval-augmented fine-tuning to multi-step RS workflow planning over procedural SOPs, with noise-robust training that leads the model to extract transferable procedural structure from noisy SOPs rather than copy them mechanically.
    \item \textbf{Multi-agent trajectory synthesis and SimRS-14k benchmark.} We introduce dual-mode user simulation paired with constraint validation in a Mock Execution Engine, producing a large-scale, constraint-validated RS tool use trajectory corpus.
\end{itemize}

\begin{figure*}[!t]
    \centering
    \includegraphics[width=1.0\linewidth]{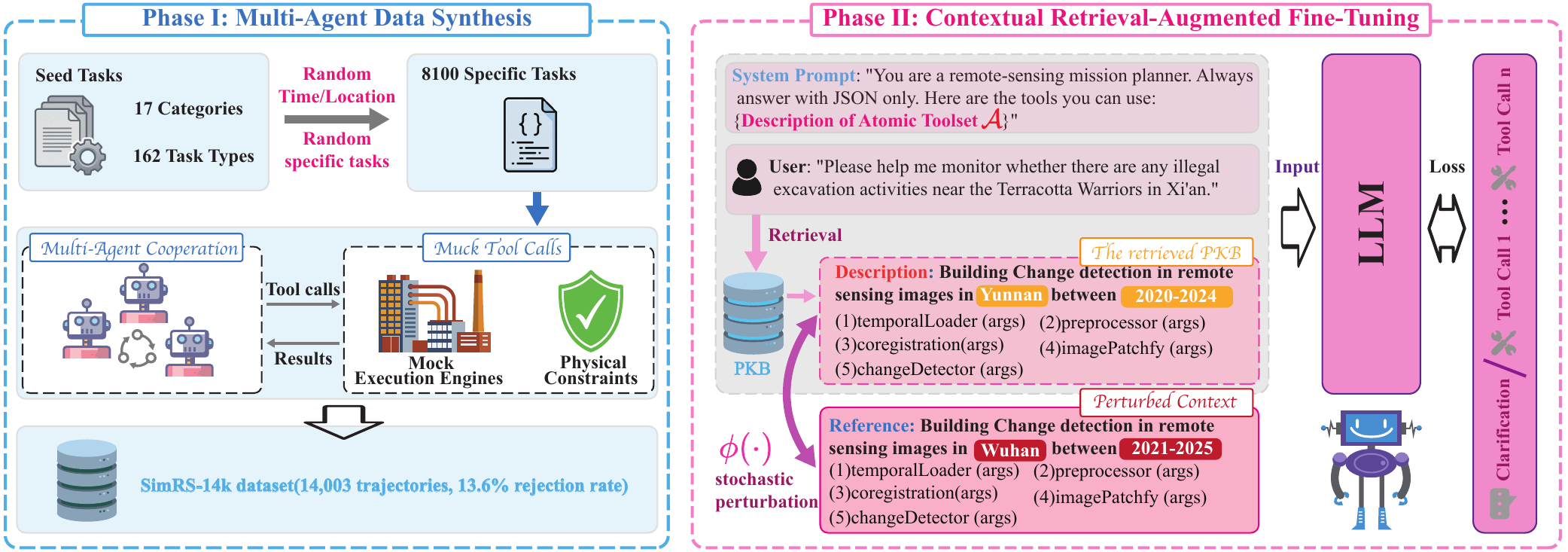}
    \caption{\textbf{Overview of the SimCRAFT framework.} Phase~I expands 162 seed tasks into the SimRS-14k corpus; Phase~II distills procedural knowledge into a 7B backbone via Contextual Retrieval-Augmented Fine-Tuning.}
    \label{fig:framework}
\end{figure*}

\section{Related Work}
\label{sec:related-work}

\subsection{Remote Sensing Agents}
RS intelligence is shifting from passive perception to active reasoning~\citep{zhang2025core,ghamisi2025geospatial}. LLM-based RS agents address the multi-step planning gap left by vision-language models such as SkyEyeGPT~\citep{zhan2025skyeyegpt}, SegEarth-R1~\citep{li2025segearth}, RingMo-Agent~\citep{hu2025ringmo}, and RSGPT~\citep{hu2025rsgpt}: EarthAgent~\citep{li2025designing} introduces a hierarchical task-abstraction mechanism, CangLing-KnowFlow~\citep{chen2025cangling} integrates a retrieval-augmented knowledge base for dynamic replanning, and frameworks such as GeoGPT~\citep{zhang2024geogpt}, RS-Agent~\citep{xu2024rs}, and GIS Copilot~\citep{akinboyewa2025gis} use multi-agent collaboration. These systems, however, all rely on GPT-4-scale closed-source LLMs for run-time orchestration. SimCRAFT instead internalizes procedural planning into the weights of a 7B-scale backbone, removing the dependence on run-time frontier models.

\subsection{Data Synthesis for Agents}

The scarcity of domain-specific planning data has driven attention to synthetic generation~\citep{li2023api}. A mature pipeline has emerged in general domains: ToolAlpaca~\citep{tang2023toolalpaca} generates multi-turn tool-use trajectories in a simulated environment, APIGen~\citep{liu2024apigen} and ToolACE~\citep{liu_toolace_2025} ensure API-call correctness through verification-based filtering and self-evolving models, and TOUCAN~\citep{xu_toucan_2025} scales the idea to millions of cross-environment interactions. Benchmarks such as AgentBench~\citep{liu2023agentbench} and GTA~\citep{Wang2024GTA,krechetova2025geobenchx} further show that reliable multi-turn tool use remains challenging. Directly transplanting these pipelines to RS leaves two gaps: existing RS corpora are dominated by shallow single-turn interactions, and RS tasks routinely involve long-horizon sequences with hard physical constraints on parameter schemas and sensor compatibility that general-domain pipelines rarely model explicitly. Our established SimRS-14k closes both gaps via dual-mode user simulation and the constraint validation of the Mock Execution Engine.

\subsection{Retrieval-Augmented Fine-Tuning}
Retrieval-Augmented Generation (RAG) typically grounds frozen LLMs with external knowledge~\citep{zhu2025knowagent}, but the retrieved documents often inject distracting context~\citep{zhang2024raft}. Retrieval-Augmented Fine-Tuning (RAFT) addresses this by bringing retrieval into the training stage, teaching the model to filter noise and reason by analogy; RAP~\citep{kagaya2024rap} and InstructRAG~\citep{wang2025instructrag} instantiate it for instruction-guided task execution, while RA-DIT~\citep{lin2023ra} and HiRAG~\citep{jiao2025hirag} perform joint generator-retriever domain adaptation. These efforts, however, almost exclusively target factual knowledge retrieval or general instruction following; \emph{agent train-time} remains an open space. While retrieval-augmented fine-tuning has targeted QA~\citep{zhang2024raft} and single-step API selection~\citep{patil2024gorilla}, our CRAFT applies it to multi-step RS workflow planning, treating RS SOPs as procedural priors with Irrelevant Context Injection and Parameter Mutation perturbations.

\section{Method}

\subsection{Task Formulation}
\label{sec:task-formulation}

We formalize the \textbf{Remote Sensing (RS) Agent} task as constrained multi-step tool calling. Given a natural-language query $q$ and an atomic-tool action space $\mathcal{A}$, the aim of the agent is to produce an executable tool-call trajectory $\tau = [(t_i, \boldsymbol{\theta}_i, r_i)]_{i=1}^{N}$, where $t_i \in \mathcal{A}$, $\boldsymbol{\theta}_i$ is the argument dictionary, and $r_i$ is the execution result. The full trajectory must satisfy a set of RS physical constraints $\mathcal{C}_{\text{RS}}$ covering sensor compatibility, data dependencies, and parameter schema legality.

The task can be naturally decomposed into three stages: \textbf{Stage~1: Intent Clarification} elicits missing parameters from $q$ through multi-turn dialogue to produce a parameter-complete intent $q^*$; \textbf{Stage~2: Procedural Retrieval} fetches the top-$k$ SOPs $\mathcal{C}_{q^*}$ from a Procedural Knowledge Base (PKB) $\mathcal{K}$ as a structural prior; and \textbf{Stage~3: Autoregressive Planning} generates each tool call from the policy $\pi_{\theta_M}$, conditioned on the system prompt $I_{\text{sys}}$, retrieved context $\tilde{\mathcal{C}}_{q^*}$, $q^*$, and execution history $\mathcal{H}_{<i}$:
\begin{equation}
\begin{aligned}
(t_i, \boldsymbol{\theta}_i) &= \pi_{\theta_M}\!\left(I_{\text{sys}},\; \tilde{\mathcal{C}}_{q^*},\; q^*,\; \mathcal{H}_{<i}\right), \\
&\quad \text{s.t.}\;\; \forall i\colon (t_i, \boldsymbol{\theta}_i) \in \mathcal{C}_{\text{RS}}.
\end{aligned}
\label{eq:task-formulation}
\end{equation}

\subsection{Framework Overview}
\label{sec:overview}

SimCRAFT is a two-phase, model-agnostic training framework, with an overview provided in Figure~\ref{fig:framework}. \textbf{Phase~I (Multi-Agent Data Synthesis)} addresses the lack of large-scale, parametrically precise, and constraint-validated training corpora in the RS domain: starting from expert-designed seed tasks, it produces the SimRS-14k corpus through dual-mode user simulation, multi-agent collaboration, and Mock Execution Engine validation. \textbf{Phase~II (CRAFT)} addresses how procedural knowledge can be written into the weights: using SimRS-14k as the supervision signal, it injects retrieved SOPs from the PKB as a contextual prior during training and applies noise perturbations that force the model to learn the structural patterns underlying SOPs rather than surface details. While the framework can be instantiated on any compact pre-trained LLM, we use Qwen-2.5-7B as the primary backbone to obtain the representative checkpoint \textbf{SimCRAFT-Qwen2.5-7B}, with backbone generality further analyzed in Section~\ref{sec:ablation-backbone-paradigm}.

\subsection{Atomic Toolset}
\label{sec:toolset}

We start our method by instantiating the abstract action space $\mathcal{A}$ from Section~\ref{sec:task-formulation} as the \textbf{Atomic Toolset}: an RS expert team designs and verifies it tool by tool, following the interface conventions of mainstream geospatial software (GDAL, SNAP, Google Earth Engine) and covering the full life cycle of RS analysis across five functional domains. Every tool is atomic to support flexible long-horizon composition, and is governed by a strictly typed JSON schema with valid value ranges to mitigate parameter hallucination. The full toolset, including JSON schemas and representative tools per domain, is provided in the appendix.

\subsection{Phase I: Multi-Agent Data Synthesis}
\label{sec:phase1}

\begin{figure}[!t]
    \centering
    \includegraphics[width=\linewidth]{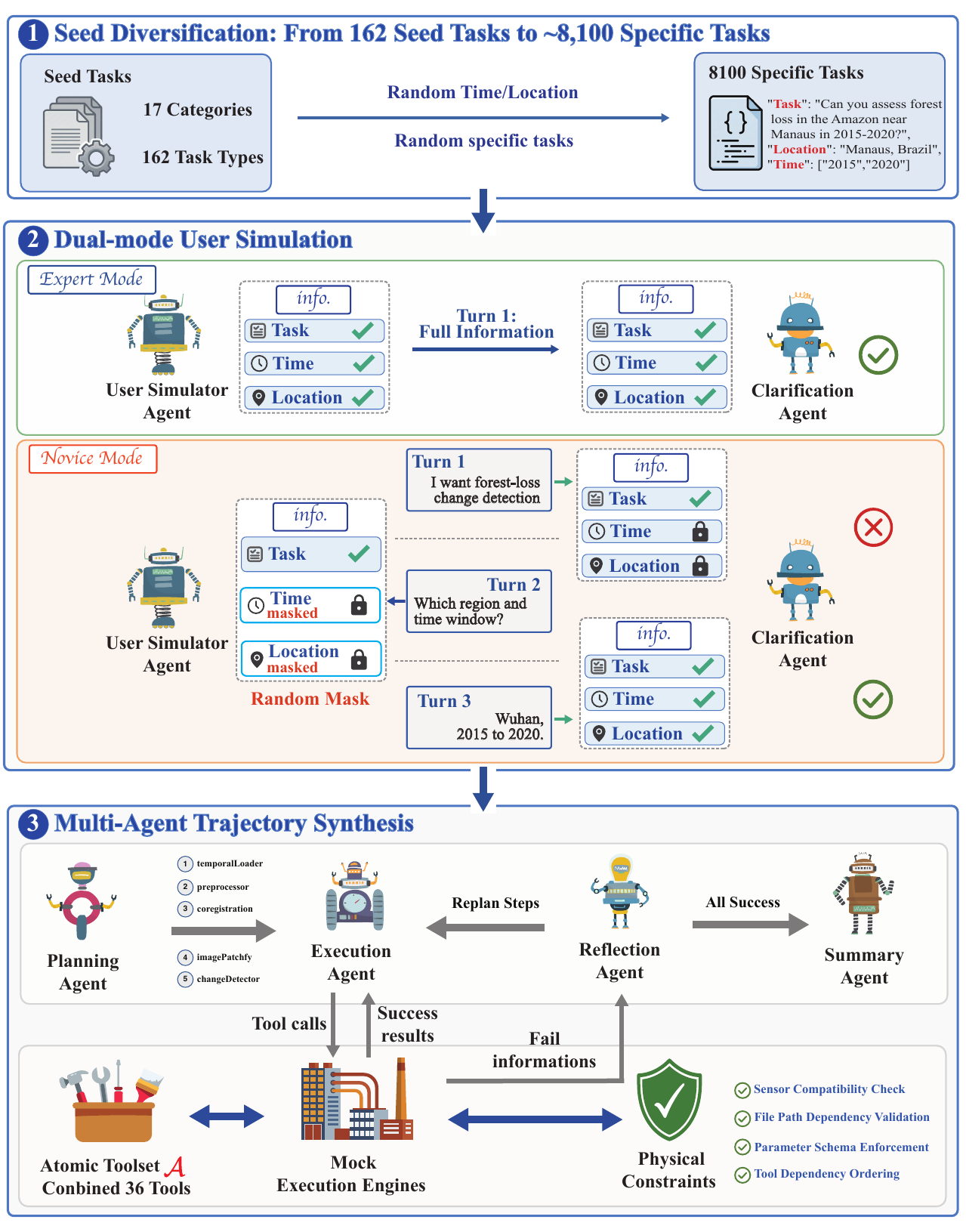}
    \caption{\textbf{Phase~I synthesis pipeline.} (a) Dual-mode user simulation, (b) multi-agent trajectory synthesis, (c) Mock Execution Engine.}
    \label{fig:synthesis}
\end{figure}

Phase~I expands the expert-designed seed tasks into the SimRS-14k corpus (Figure~\ref{fig:synthesis}): \textbf{Dual-mode User Simulation} diversifies user intents, \textbf{Multi-Agent Trajectory Synthesis} expands each intent into a multi-step tool-call trajectory, and the \textbf{Mock Execution Engine} applies three validation checks at every step. Consequently, downstream training admits only trajectories that pass all these checks.

\subsubsection{Dual-mode User Simulation}
\label{sec:dualmode}

Existing agent training corpora assume that users state every critical parameter in a single query, yet RS analysts routinely omit constraints such as the time window, sensor, or spatial extent on which downstream tools strictly depend, and an agent filling these gaps with hallucinated defaults silently inherits the errors into the training target. We therefore introduce two collaborating agents: the \textbf{User Simulation Agent} switches between Expert Mode (stating all parameters at once) and Novice Mode (random parameter masking that leaves an explicit information gap), and the \textbf{Clarification Agent} maintains a ledger of critical information and initiates a targeted follow-up whenever a slot is missing. The dialogue terminates only after all critical information is aligned, so the distilled small model learns both when to ask and when to act.

\subsubsection{Multi-Agent Trajectory Synthesis}

Letting a single LLM freely generate multi-step tool-call trajectories leads to two typical errors: fabricated tool-call orderings and violations of inter-tool physical dependencies. Furthermore, on long-horizon tasks, a single agent tends to fall into unrecoverable local errors, ultimately producing formally legal but semantically invalid \emph{hallucinated trajectories}. To overcome these limitations, we decompose single-agent self-synthesis into a system of four role-complementary agents. The Atomic Toolset description from Section~\ref{sec:toolset} acts as the system prompt to constrain every agent's action space. Specifically, the \textbf{Planning Agent} drafts an initial high-level roadmap, and the \textbf{Execution Agent} produces a concrete atomic tool action $a_t$ based on the current context. If an execution fails, the \textbf{Reflection Agent} activates to perform dynamic error recovery, while the \textbf{Summary Agent} aggregates the final result upon trajectory completion. These agents coordinate through a shared state machine rather than direct message passing. By letting the current state determine the active agent at each step, the system ensures the Reflection Agent can intercept and rewrite local errors before they propagate.

\subsubsection{Mock Execution Engine Check}

Even with four-agent collaboration, the LLM can still emit silent errors such as schema violations, fabricated file paths, or sensor-compatibility violations, which poison the training target if uncaught, while plugging the pipeline into real geospatial software is prohibitively expensive since petabyte-scale imagery cannot support large-scale synthesis. We therefore design a lightweight \textbf{Mock Execution Engine} $\mathcal{E}$ that applies three independent validation checks to each tool call $(t_i, \boldsymbol{\theta}_i)$. First, the \textbf{schema check} validates argument types and value ranges. Second, the \textbf{dependency check}, implemented by a \textbf{Dynamic Path Registry} $\mathcal{R}$, rejects any call referencing an unregistered path with a \texttt{PathNotFoundError}. Third, the \textbf{compatibility check} rejects a call that violates sensor or tool compatibility, such as a spectral index requiring a near-infrared band on a product that lacks it. Beyond these validation checks, an \textbf{error injection} step emits recoverable errors (e.g., \texttt{CloudCoverExceeded}) with a small probability, forcing the model to learn the ``try, fail, correct'' recovery pattern. Successful results are returned to the Execution Agent to advance the state, while failure information is routed to the Reflection Agent to trigger replanning. Across 16,200 candidate trajectories, the three checks jointly yield a \textbf{rejection rate of 13.6\%}, leaving 14,003 trajectories that constitute \textbf{SimRS-14k}.

Beyond the automatic checks, four RS experts manually inspect a stratified random sample of 2800 trajectories, about 20\% of SimRS-14k, stratified by seed task and trajectory length. For each sampled instance they judge both whether the full workflow executes correctly and whether it answers the query. In total, 2787 of the 2800 instances, or 99.5\%, pass. We keep the 13 flagged instances in the corpus rather than removing them, treating them as low-rate residual noise consistent with the noise-robust objective of CRAFT.

\begin{table*}[t!]
\centering
\caption{\textbf{Main comparison results on the SimRS-14k test split.} \textbf{Bold} = best, \underline{underline} = second-best within each metric. ``Open'' marks open-weights backbones.}
\label{tab:main_results}
\resizebox{\textwidth}{!}{%
\begin{tabular}{llccccccc}
\toprule
\textbf{Category} & \textbf{Model} & \textbf{Open} & \textbf{Size} & \textbf{PSR (\%)} & \textbf{TSA (\%)} & \textbf{AEM (\%)} & \textbf{Hal.Rate (\%)} $\downarrow$ & \textbf{Fmt.Err (\%)} $\downarrow$ \\
\midrule
\multirow{10}{*}{Generalist LLMs}
& Llama-3-8B-Instruct & $\checkmark$ & 8B & 42.5 & 58.2 & 35.1 & 15.4 & 32.6 \\
& Qwen-2.5-7B-Instruct & $\checkmark$ & 7B & 46.2 & 62.5 & 39.8 & 12.1 & 26.4 \\
& Qwen-2.5-14B-Instruct & $\checkmark$ & 14B & 58.6 & 71.8 & 48.9 & 7.5 & 18.5 \\
& Qwen-2.5-32B-Instruct & $\checkmark$ & 32B & 68.4 & 78.1 & 55.4 & 4.5 & 12.8 \\
& Qwen-2.5-72B-Instruct & $\checkmark$ & 72B & 74.6 & 82.4 & 62.8 & 2.7 & 8.1 \\
& DeepSeek-V3 & $\checkmark$ & 671B & 76.9 & 84.5 & 66.4 & 3.2 & 6.7 \\
& DeepSeek-R1 & $\checkmark$ & 671B & 77.5 & 85.8 & 68.1 & 2.8 & 6.6 \\
& GPT-4 & $\times$ & GPT-4 & 81.1 & 89.1 & 72.3 & 1.2 & 5.1 \\
& GPT-5 & $\times$ & GPT-5 & 83.8 & 90.6 & 74.5 & \underline{0.6} & \underline{4.5} \\
& Gemini-3-Flash & $\times$ & Gemini-3-Flash & 84.1 & 91.2 & 75.8 & 0.9 & \textbf{4.2} \\
\midrule
\multirow{6}{*}{General Agents}
& ReAct (Llama-3-8B) & $\checkmark$ & 8B & 52.1 & 65.3 & 41.9 & 8.5 & 22.5 \\
& Reflexion (Llama-3-8B) & $\checkmark$ & 8B & 58.4 & 70.1 & 48.5 & 6.2 & 16.8 \\
& AFlow (Llama-3-8B) & $\checkmark$ & 8B & 61.9 & 72.8 & 51.2 & 5.8 & 14.2 \\
& ReAct (GPT-4) & $\times$ & GPT-4 & 72.4 & 82.3 & 61.2 & 2.4 & 8.9 \\
& Reflexion (GPT-4) & $\times$ & GPT-4 & 76.8 & 85.4 & 65.6 & 2.2 & 6.8 \\
& AFlow (GPT-4) & $\times$ & GPT-4 & 79.2 & 87.6 & 68.4 & 1.1 & 5.5 \\
\midrule
\multirow{4}{*}{RS Agents}
& EarthAgent (DeepSeek-R1) & $\checkmark$ & 671B & 80.6 & 87.2 & 70.4 & 1.6 & 6.2 \\
& CangLing-KnowFlow (DeepSeek-R1) & $\checkmark$ & 671B & 81.8 & 88.4 & 71.6 & 1.4 & 5.8 \\
& EarthAgent (GPT-4) & $\times$ & GPT-4 & \underline{86.2} & 91.8 & 76.8 & 0.7 & 4.6 \\
& CangLing-KnowFlow (GPT-4) & $\times$ & GPT-4 & \textbf{87.5} & \underline{92.1} & \underline{77.2} & 0.8 & 5.2 \\
\midrule
\rowcolor{gray!10} \textbf{Ours} & \textbf{SimCRAFT-Qwen2.5-7B} & $\checkmark$ & \textbf{7B} & 83.2 & \textbf{93.5} & \textbf{79.6} & \textbf{0.5} & 5.4 \\
\bottomrule
\end{tabular}%
}
\end{table*}

\subsection{Phase II: CRAFT}
\label{sec:phase2}

Phase~II distills SimRS-14k together with the SOPs in the PKB into the weights of a 7B-scale pre-trained LLM (see the Phase~II part of Figure~\ref{fig:framework}). CRAFT realizes this through two components: \textbf{PKB Retrieval} fetches top-$k$ SOPs as a procedural prior at both training and inference time, and \textbf{Noise-Robust Training} perturbs the retrieved context so that the model learns structure rather than surface details, with the same frozen retriever reused at deployment to produce executable trajectories.

\subsubsection{PKB Retrieval}
\label{sec:pkb-retrieval}

Directly applying end-to-end SFT on SimRS-14k causes a 7B model to memorize training instances, severely degrading its generalization to queries beyond the seed tasks. On the other hand, relying solely on inference-time SOP injection into a frozen model introduces redundant signals that disrupt planning, primarily because small models lack the capacity to extract procedural structures from heterogeneous contexts. We therefore construct a Procedural Knowledge Base $\mathcal{K} = \{(q_i,\tau_i)\}_{i=1}^{M}$ that pairs each task query $q_i$ with its expert-validated SOP $\tau_i$, initialized from the expert workflows of CangLing-KnowFlow~\citep{chen2025cangling}. The retriever is a frozen dual-encoder $E(\cdot)$, shared between training and inference, that selects from $\mathcal{K}$ the top-$k$ trajectories whose cosine similarity with the query $x$ exceeds a threshold $\delta$ as the procedural prior:
\begin{equation}
\begin{aligned}
\mathcal{C}_x &= \mathop{\text{Top-}k}_{\tau_i}\!\left(\{\tau_i \mid (q_i, \tau_i) \in \mathcal{K}\}\right), \\
&\quad \text{s.t. } \text{sim}(E(x), E(q_i)) \ge \delta.
\end{aligned}
\label{eq:pkb-retrieval}
\end{equation}

\subsubsection{Noise-Robust Training}
\label{sec:noise-robust}

Concatenating the retrieved $\mathcal{C}_x$ directly into the prompt under standard RAFT training causes the model to take the shortcut of copying $\mathcal{C}_x$, so that any drop in retrieval quality at inference time leads it to follow the flawed template and the whole trajectory collapses. We therefore apply a stochastic perturbation $\phi(\cdot)$ to $\mathcal{C}_x$ at training time: (i) \textbf{Irrelevant Context Injection} ($\phi_{ICI}$) replaces a reference trajectory with a semantically unrelated one $\tau_{\text{noise}}$ with a given probability; and (ii) \textbf{Parameter Mutation} ($\phi_{PM}$) randomly rewrites parameters within a valid reference, e.g., changing \texttt{Sentinel-2} to \texttt{Landsat-8}, forcing the model to re-verify physical constraints. Given a query $x$, target trajectory $y$, and perturbed context $\tilde{\mathcal{C}}_x = \phi(\mathcal{C}_x)$, the CRAFT training objective is
\begin{equation}
\mathcal{L}_{\text{CRAFT}} = -\sum_{t=1}^{T} \log P_\theta\!\left(y_t \mid x,\; \tilde{\mathcal{C}}_x,\; y_{<t}\right).
\label{eq:craft-loss}
\end{equation}

Because the supervision $y$ is always a constraint-validated correct trajectory while $\tilde{\mathcal{C}}_x$ may contain incorrect fragments, the model must extract transferable procedural structures from the noisy context and actively override incorrect details. This process essentially serves as a context-denoising objective, teaching the model when to ignore retrieved content during the training phase. At deployment, the same frozen retriever fetches top-$k$ SOPs for $q^*$ with no perturbation applied, and SimCRAFT-Qwen2.5-7B autoregressively generates tool calls following Equation~\ref{eq:task-formulation} until \texttt{Terminate}.

\section{Experiments}

\subsection{Experimental Setup}
\label{sec:exp-setup}

\paragraph{Datasets and implementation.} The Phase-I pipeline (Section~\ref{sec:phase1}) expands 162 seed tasks into SimRS-14k. After validation via the Mock Execution Engine, we retain 14,003 multi-turn tool-call trajectories. We hold out 500 testing trajectories via stratified sampling to ensure strictly no overlap with the training set in geographic, temporal, or sensor parameters, leaving the remaining 13,503 for Phase-II CRAFT fine-tuning. To ensure rigorous evaluation, we perform zero-shot testing on KnowFlow-Bench~\citep{chen2025cangling}, a collection of 324 expert-annotated real-world RS tasks completely isolated from our training pipeline. For implementation, SimCRAFT-Qwen2.5-7B is built upon Qwen-2.5-7B-Instruct~\citep{yang2024qwen2} using LoRA~\citep{hu2022lora} ($r=64$, $\alpha=128$, dropout 0.05). We train for 3 epochs using AdamW (learning rate $2\times 10^{-5}$) and a cosine schedule. Both the error-injection and CRAFT perturbation probabilities are set to $\varepsilon=0.15$. During retrieval, we apply a threshold of $\delta=0.6$ and $k=2$. All experiments run on 8 NVIDIA A800 GPUs. We report 95\% confidence intervals for SimCRAFT-Qwen2.5-7B from 1000 bootstrap resamples of the test set in Appendix~\ref{app:ci}.

The PKB provides retrieval templates and both the PKB and KnowFlow-Bench derive from CangLing-KnowFlow, so we additionally evaluate on the independently sourced ThinkGeo benchmark (Section~\ref{sec:thinkgeo}) to rule out any shared-provenance effect.

\subsection{Baselines}
\label{sec:baselines}

\paragraph{Baselines.} Table~\ref{tab:main_results} compares against three categories: (i) \textbf{Generalist LLMs}: open-weights Llama-3-8B-Instruct~\citep{dubey2024llama}, the Qwen-2.5 Instruct family from 7B to 72B~\citep{yang2024qwen2}, DeepSeek-V3~\citep{liu2024deepseek}, DeepSeek-R1~\citep{guo2025deepseek}, and proprietary GPT-4~\citep{achiam2023gpt}, GPT-5, Gemini-3-Flash; (ii) \textbf{General Agents}: ReAct~\citep{yao2023react}, Reflexion~\citep{shinn2023reflexion}, AFlow~\citep{zhang2024aflow}, each instantiated on both Llama-3-8B and GPT-4; (iii) \textbf{RS Agents}: EarthAgent~\citep{li2025designing} (hierarchical planning) and CangLing-KnowFlow~\citep{chen2025cangling} (static RAG), each instantiated on DeepSeek-R1 and GPT-4. A same-scale \emph{General Agents + Inf-RAG (no CRAFT)} configuration is reserved as an ablation variant in Section~\ref{sec:ablations}.

\subsection{Evaluation Metrics}
\label{sec:metrics}

We adopt a five-metric protocol covering correctness, planning quality, and robustness. \textbf{Planning Success Rate (PSR)} is the primary metric, measuring the proportion of tasks for which the generated toolchain is both executable and logically aligned with the ground truth under the longest common subsequence~\citep{michelakis2025core}. \textbf{Tool Selection Accuracy (TSA)} and \textbf{Argument Exact Match (AEM)} decompose planning quality into tool-name hit rate~\citep{li2025designing} and the percentage of arguments that strictly match the schema~\citep{shabbir2025thinkgeo}. \textbf{Hallucination Rate (Hal.Rate)} and \textbf{Format Error Rate (Fmt.Err)} measure robustness as the frequencies of unregistered-path / non-existent-tool references and invalid JSON outputs~\citep{shabbir2025thinkgeo}, respectively (lower-is-better). Since PSR and AEM use strict matching, the reported values are conservative lower bounds, whereas TSA depends only on tool-name selection and is invariant to tool ordering and argument values.

\subsection{Main Results}
\label{sec:main-results}
Table~\ref{tab:main_results} shows the 7B-parameter SimCRAFT-Qwen2.5-7B demonstrates highly competitive performance on SimRS-14k. It matches top-tier proprietary models (GPT-5, Gemini-3-Flash) and surpasses the 671B DeepSeek-R1 in PSR (83.2\%), while dominating all generalist LLMs in TSA and AEM. Furthermore, it outperforms GPT-4-driven general agents by a +4.0\% PSR margin and significantly minimizes hallucinations to 0.5\%. Although GPT-4-based RS agents still hold a slight PSR advantage due to massive runtime orchestration, SimCRAFT decisively overtakes them in execution precision and hallucination control.

\subsection{Generalization to KnowFlow-Bench}
\label{sec:knowflow-bench}

Table~\ref{tab:knowflow_results} reports the zero-fine-tuning cross-benchmark comparison. SimCRAFT-Qwen2.5-7B attains 79.4\% PSR, surpassing the same-backbone Inf-RAG baseline by 29.2\%, trailing the $100{\times}$-larger GPT-4~+~Inf-RAG by only 3.2\%, and exhibiting a cross-benchmark drop ($-3.8\%$) roughly $3{\times}$ smaller than vanilla SFT (supervised fine-tuning on SimRS-14k trajectories without PKB context, $-10.8\%$). This indicates that CRAFT learns transferable tool-use structure rather than surface templates of the 162 seed tasks.

\begin{table}[t]
\centering
\small
\renewcommand{\arraystretch}{1.15}
\begin{tabular*}{\columnwidth}{@{\extracolsep{\fill}}lccc@{}}
\toprule
\textbf{Method} & \textbf{Params} & \textbf{PSR} & \textbf{AEM} \\
\midrule
Llama-3-8B~+~Inf-RAG & 8B & 46.3 & 51.2 \\
Mistral-7B~+~Inf-RAG & 7B & 47.8 & 53.5 \\
Qwen-2.5-7B~+~Inf-RAG & 7B & 50.2 & 55.1 \\
\midrule
Vanilla SFT-Qwen2.5-7B & 7B & 62.8 & 65.4 \\
\rowcolor{gray!10} \textbf{SimCRAFT-Qwen2.5-7B} & \textbf{7B} & 79.4 & 75.2 \\
\midrule
GPT-4~+~Inf-RAG & ${\sim}1.8$T & 82.6 & 78.3 \\
CangLing-KnowFlow (GPT-4) & ${\sim}1.8$T & \textbf{83.7} & \textbf{80.1} \\
\bottomrule
\end{tabular*}
\caption{\textbf{Zero-fine-tuning evaluation on KnowFlow-Bench} (324 expert-annotated tasks), all numbers in \%.}
\label{tab:knowflow_results}
\end{table}

\subsection{Generalization to ThinkGeo}
\label{sec:thinkgeo}

To test generalization beyond our own data, we evaluate SimCRAFT-Qwen2.5-7B with no fine-tuning on ThinkGeo~\citep{shabbir2025thinkgeo}, an independently constructed RS agent benchmark of 486 tasks and 1778 expert-verified steps over real Earth-observation imagery, with its own 14-tool action space and no shared provenance with our synthesis pipeline or CangLing-KnowFlow. We adopt its 14 tools directly as the action space and score predictions against its expert-verified references under our PSR, TSA, AEM, and Hal.Rate protocol. At inference SimCRAFT follows its standard pipeline, retrieving the most similar SOPs from the PKB before planning, without any adaptation to ThinkGeo. As shown in Table~\ref{tab:thinkgeo_results}, the same-size baselines drop sharply on this unseen tool space while SimCRAFT-Qwen2.5-7B stays close to GPT-4, indicating that the planning ability transfers beyond our synthetic distribution and KnowFlow-Bench.

\begin{table}[t]
\centering
\small
\renewcommand{\arraystretch}{1.15}
\setlength{\tabcolsep}{3pt}
\begin{tabular*}{\columnwidth}{@{\extracolsep{\fill}}lcccc@{}}
\toprule
\textbf{Model} & \textbf{PSR} & \textbf{TSA} & \textbf{AEM} & \textbf{Hal.Rate} $\downarrow$ \\
\midrule
GPT-4 (reference) & 61.8 & 67.4 & 34.9 & 2.3 \\
Qwen-2.5-7B (zero-shot) & 29.5 & 51.0 & 20.1 & 18.6 \\
Llama-3-8B (zero-shot) & 24.8 & 37.3 & 13.7 & 22.4 \\
\rowcolor{gray!10} \textbf{SimCRAFT-Qwen2.5-7B} & 58.3 & 66.5 & 32.8 & 3.9 \\
\bottomrule
\end{tabular*}
\caption{\textbf{Zero-fine-tuning evaluation on ThinkGeo}~\citep{shabbir2025thinkgeo} (486 tasks), all numbers in \%. SimCRAFT-Qwen2.5-7B uses no fine-tuning on ThinkGeo.}
\label{tab:thinkgeo_results}
\end{table}

\subsection{Ablation Studies}
\label{sec:ablations}

\begin{figure*}[t]
    \centering
    \includegraphics[width=1.0\linewidth]{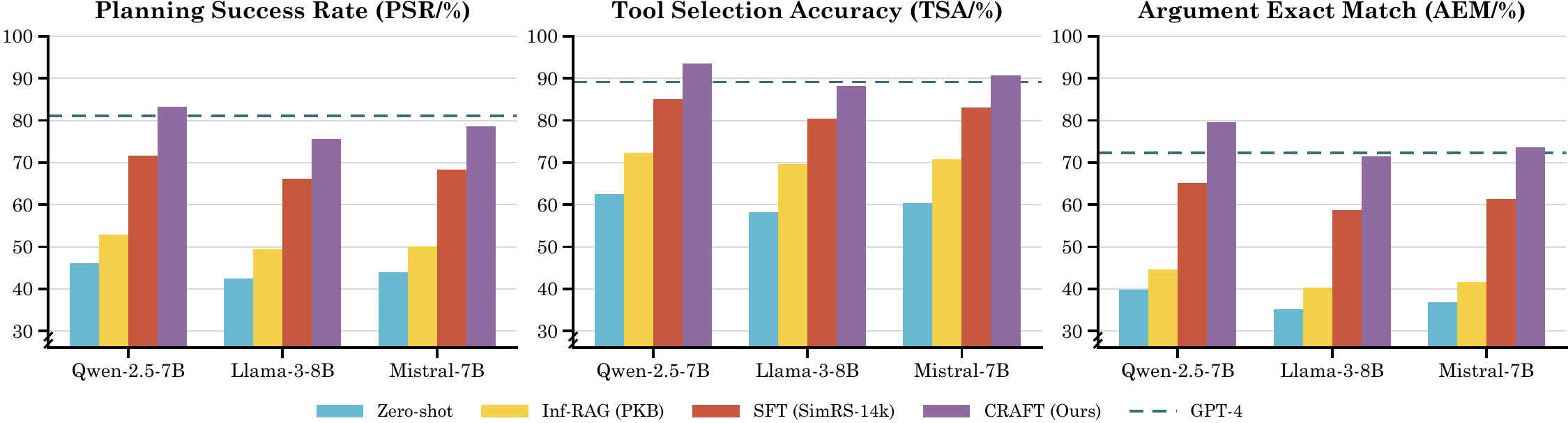}
    \caption{\textbf{Joint backbone $\times$ training-paradigm ablation.} Three 7B-scale backbones each evaluated under four paradigms (Zero-Shot, Inf-RAG, vanilla SFT, CRAFT).}
    \label{fig:backbone-paradigm}
\end{figure*}

\paragraph{Backbone \& Training-Paradigm.}
\label{sec:ablation-backbone-paradigm}
We evaluate three 7B-scale backbones (Qwen-2.5-7B, Llama-3-8B, Mistral-7B~\citep{jiang2023mistral}) across four paradigms: Zero-Shot, Inf-RAG (inference-time PKB retrieval into the prompt, no fine-tuning), vanilla SFT, and CRAFT (Figure~\ref{fig:backbone-paradigm}). Across all backbones, CRAFT consistently yields a $+9.6\%$ to $+11.6\%$ PSR improvement over vanilla SFT without any paradigm rank-inversion. Furthermore, the 7.5\% PSR range in the CRAFT column is comparable to the 3.7\% inherent capability gap observed in the Zero-Shot column, supporting our model-agnostic positioning. Extending the study to a more recent 2025 backbone, Qwen3-8B~\citep{qwen3}, further raises CRAFT to 84.7\% PSR and leaves all conclusions unchanged, confirming that the approach tracks backbone progress.

\begin{figure}[t]
    \centering
    \includegraphics[width=1.0\linewidth]{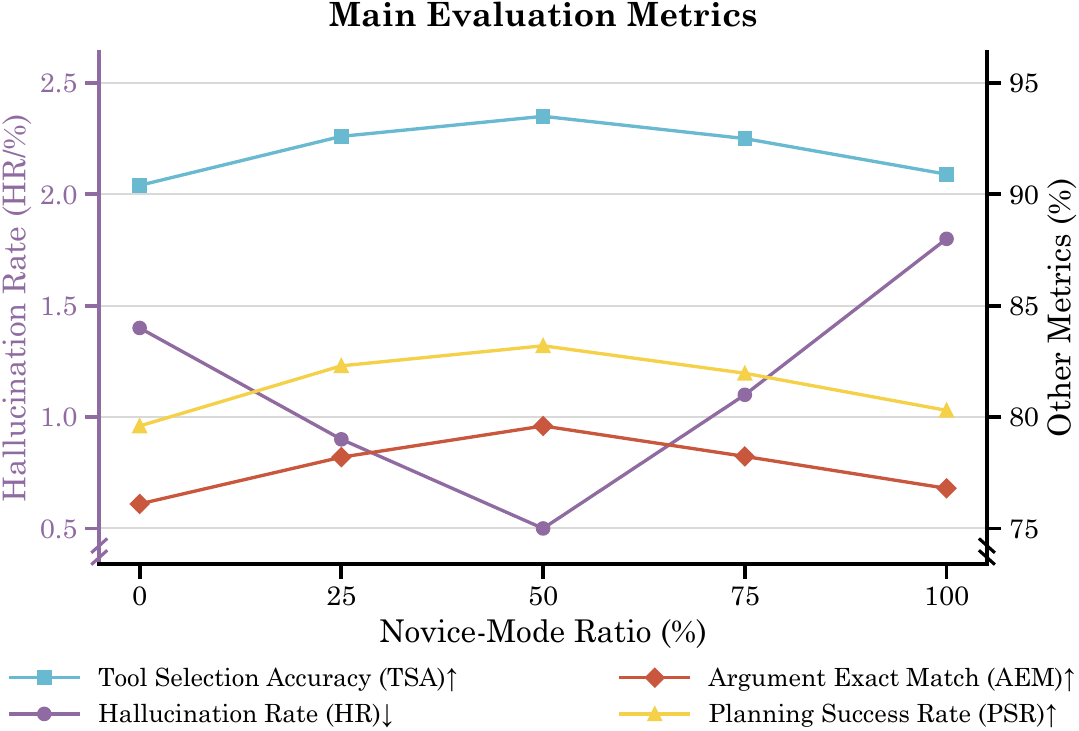}
    \caption{\textbf{Clarification ratio ablation.} PSR, TSA, AEM (right y axis), and Hal.Rate (left y axis) across five novice fractions from 0\% to 100\%.}
    \label{fig:clarification-ratio}
\end{figure}

\paragraph{Novice Mode Ratio.}
\label{sec:ablation-clarification}
Maintaining a fixed total corpus size of 13,974 trajectories, we evaluate the novice fraction across five distinct ratios (Figure~\ref{fig:clarification-ratio}). We observe that all four evaluation metrics simultaneously reach their optimal values at an even 50/50 split, achieving a PSR of 83.2\%, TSA of 93.5\%, AEM of 79.6\%, and a Hallucination Rate as low as 0.5\%. Conversely, models trained on either extreme ratio yield a PSR below 80.3\%. This performance degradation demonstrates that the efficacy of our dual-mode approach stems from the synergistic coexistence of both distributions, rather than the exclusive scaling of a single data type.

\paragraph{Noise-Robust Training.}
\label{sec:ablation-noise}
Varying the perturbation scale $\varepsilon \in \{0, 0.05, 0.10, 0.15, 0.30, 0.50\}$ (Table~\ref{tab:noise-eps}) yields a peak PSR of 83.2\% at $\varepsilon=0.15$, whereas excessive noise ($\varepsilon=0.50$) degrades performance below the unperturbed baseline, explicitly refuting the assumption of monotonic improvements. At this optimal setting (Table~\ref{tab:noise-comp}), the combined 4.6\% gain from Irrelevant Context Injection (+3.1\%) and Parameter Mutation (+1.8\%) closely approximates their linear sum (4.9\%), indicating these mechanisms operate as independent factors.

\begin{table}[t]
\centering
\setlength{\tabcolsep}{4pt}
\begin{tabular*}{\columnwidth}{@{\extracolsep{\fill}}lcccccc@{}}
\toprule
$\varepsilon$ & $0.00$ & $0.05$ & $0.10$ & $\mathbf{0.15}$ & $0.30$ & $0.50$ \\
\midrule
PSR (\%) & $78.6$ & $80.7$ & $82.4$ & $\mathbf{83.2}$ & $81.1$ & $76.9$ \\
\bottomrule
\end{tabular*}
\caption{\textbf{Perturbation probability $\varepsilon$ sweep} on SimRS-14k with both perturbation types active with backbone fixed to Qwen-2.5-7B.}
\label{tab:noise-eps}
\end{table}

\begin{table}[t]
\centering
\begin{tabular*}{\columnwidth}{@{\extracolsep{\fill}}lcc@{}}
\toprule
Configuration & PSR (\%) & AEM (\%) \\
\midrule
vanilla RAFT ($\phi$ off) & $78.6$ & $73.4$ \\
\quad + $\phi_{ICI}$ only & $81.7$ & $75.3$ \\
\quad + $\phi_{PM}$ only & $80.4$ & $78.2$ \\
\rowcolor{gray!10} \textbf{\quad + both (full)} & $\mathbf{83.2}$ & $\mathbf{79.6}$ \\
\bottomrule
\end{tabular*}
\caption{\textbf{Component breakdown of the two perturbation types} at the default $\varepsilon{=}0.15$. ICI: Irrelevant Context Injection; PM: Parameter Mutation.}
\label{tab:noise-comp}
\end{table}

\subsection{In-Depth Analysis}
\label{sec:in-depth-analysis}

\begin{figure}[t]
    \centering
    \includegraphics[width=1.0\linewidth]{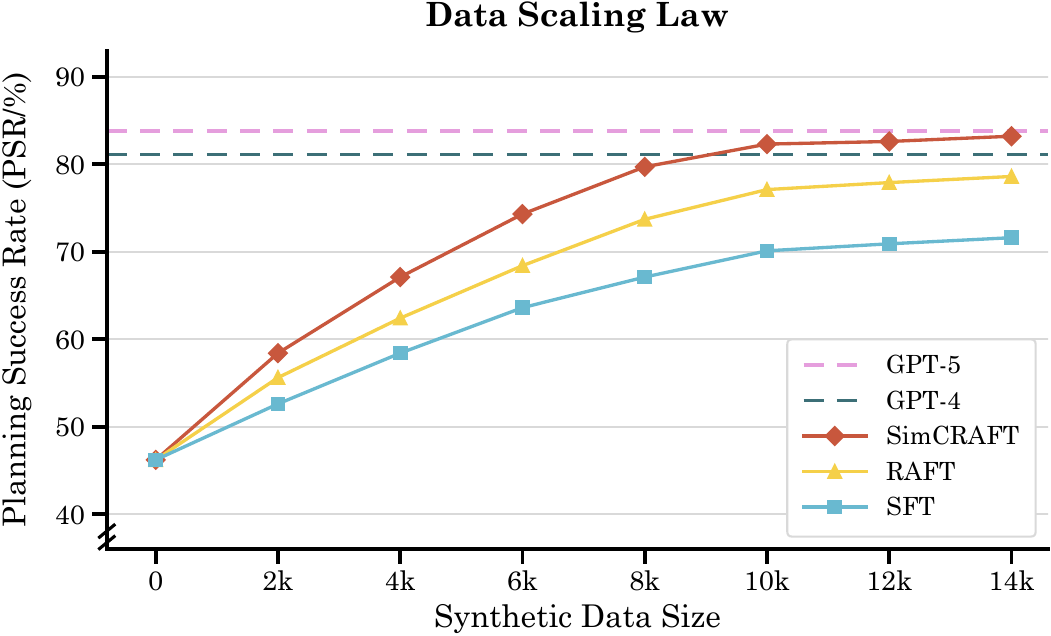}
    \caption{\textbf{Data scaling law.} PSR vs.\ SimRS-14k corpus size, with GPT-4 / GPT-5 zero-shot reference lines.}
    \label{fig:scaling}
\end{figure}

\textbf{Data Scaling Law.} Figure~\ref{fig:scaling} shows that PSR scales near log-linearly with the size of SimRS-14k: SimCRAFT (Full) matches the 81.1\% PSR of GPT-4 (zero-shot) at around 9k samples and reaches 83.2\% at 14k, close to GPT-5's 83.8\%.

\begin{figure}[t!]
    \centering
    \includegraphics[width=1.0\linewidth]{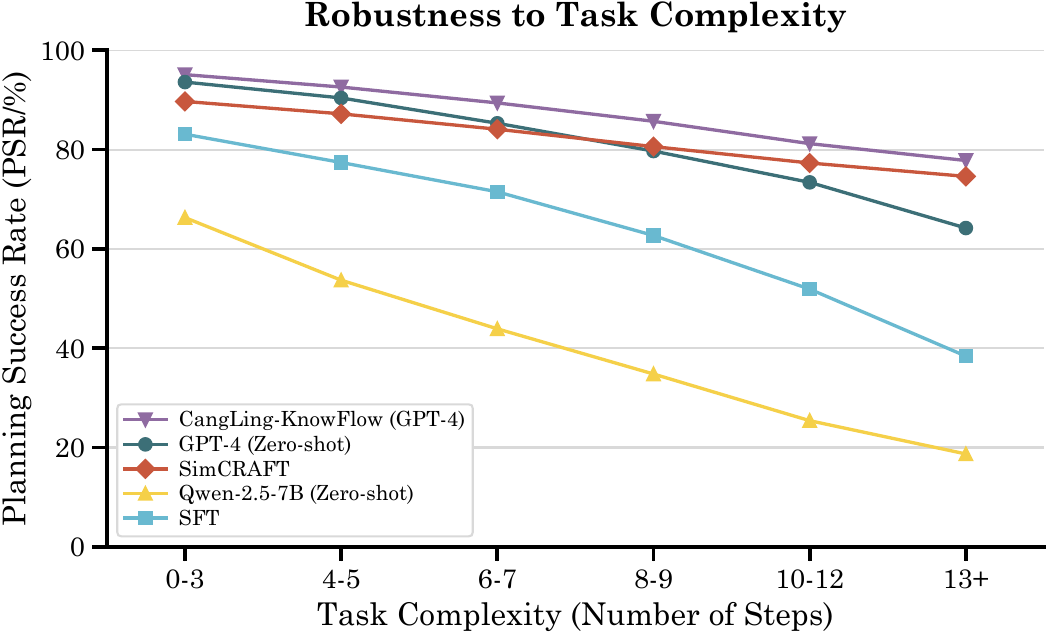}
    \caption{\textbf{Long-horizon robustness.} PSR by trajectory-length bucket.}
    \label{fig:longhorizon}
\end{figure}
\textbf{Long-Horizon Robustness.} While baselines suffer severe degradation on extended trajectories (e.g., zero-shot Qwen-2.5-7B plummets from 66.3\% to 18.7\%), SimCRAFT exhibits remarkable stability across all lengths (Figure~\ref{fig:longhorizon}), maintaining a PSR of at least 74.6\% and decisively surpassing zero-shot GPT-4 on tasks exceeding 8 steps.

\section{Conclusion}
\label{sec:conclusion}

We present \textbf{SimCRAFT}, a model-agnostic training framework that distills RS-agent procedural planning into a 7B open-source model. By combining the constraint-validated SimRS-14k corpus with the noise-robust CRAFT training paradigm, SimCRAFT-Qwen2.5-7B matches GPT-4-level RS agents on SimRS-14k (83.2\% PSR, 93.5\% TSA, 0.5\% Hal.Rate) and generalizes to KnowFlow-Bench at 79.4\% PSR without any additional fine-tuning. The results demonstrate that constraint-validated synthetic data combined with a context-denoising training objective can align a $100{\times}$ smaller open-source model with closed-source frontier LLMs on RS procedural planning.
\section*{Limitations}
\label{sec:limitations}

The Mock Execution Engine validates only schema, dependency, and sensor compatibility rather than invoking GDAL/SNAP/GEE on real imagery, so end-to-end physical-accuracy evaluation of downstream science products (e.g., change-detection Kappa) remains future work. The Atomic Toolset and PKB are restricted to RS-specific tools, leaving cross-domain transfer (e.g., medical-imaging or bioinformatics pipelines) requiring a domain-specific reconstruction. Our metrics PSR, TSA, and AEM assess workflow-planning quality and do not measure the correctness of downstream products such as change maps, classification maps, or statistics, which we scope out as an explicit limitation.

\bibliography{simcraft}

@article{li2025segearth,
  title={Seg{E}arth-{R}1: Geospatial pixel reasoning via large language model},
  author={Li, Kaiyu and Xin, Zepeng and Pang, Li and Pang, Chao and Deng, Yupeng and Yao, Jing and Xia, Guisong and Meng, Deyu and Wang, Zhi and Cao, Xiangyong},
  journal={arXiv preprint arXiv:2504.09644},
  year={2025}
}

@article{ghamisi2025geospatial,
  title={Geospatial Foundation Models to Enable Progress on Sustainable Development Goals},
  author={Ghamisi, Pedram and Yu, Weikang and Zhang, Xiaokang and Rizaldy, Aldino and Wang, Jian and Zhou, Chufeng and Gloaguen, Richard and Camps-Valls, Gustau},
  journal={arXiv preprint arXiv:2505.24528},
  year={2025}
}

@inproceedings{li2023api,
  title={{API-Bank}: A Comprehensive Benchmark for Tool-Augmented {LLM}s},
  author={Li, Minghao and Zhao, Yingxiu and Yu, Bowen and Song, Feifan and Li, Hangyu and Yu, Haiyang and Li, Zhoujun and Huang, Fei and Li, Yongbin},
  booktitle={Proceedings of the 2023 Conference on Empirical Methods in Natural Language Processing},
  pages={3102--3116},
  year={2023}
}

@inproceedings{liu2023agentbench,
  title={{AgentBench}: Evaluating {{LLM}}s as agents},
  author={Liu, Xiao and Yu, Hao and Zhang, Hanchen and Xu, Yifan and Lei, Xuanyu and Lai, Hanyu and Gu, Yu and Ding, Hangliang and Men, Kaiwen and Yang, Kejuan and others},
  booktitle={The Twelfth International Conference on Learning Representations},
  year={2024}
}

@article{shinn2023reflexion,
  title={Reflexion: Language agents with verbal reinforcement learning},
  author={Shinn, Noah and Cassano, Federico and Gopinath, Ashwin and Narasimhan, Karthik and Yao, Shunyu},
  journal={Advances in Neural Information Processing Systems},
  volume={36},
  pages={8634--8652},
  year={2023}
}

@article{tang2023toolalpaca,
  title={{ToolAlpaca}: Generalized tool learning for language models with 3000 simulated cases},
  author={Tang, Qiaoyu and Deng, Ziliang and Lin, Hongyu and Han, Xianpei and Liang, Qiao and Cao, Boxi and Sun, Le},
  journal={arXiv preprint arXiv:2306.05301},
  year={2023}
}

@inproceedings{yao2023react,
  title={Re{A}ct: Synergizing reasoning and acting in language models},
  author={Yao, Shunyu and Zhao, Jeffrey and Yu, Dian and Du, Nan and Shafran, Izhak and Narasimhan, Karthik and Cao, Yuan},
  booktitle={International Conference on Learning Representations (ICLR)},
  pages={1--18},
  year={2023}
}

@article{xu2024rs,
  title={RS-{A}gent: Automating Remote Sensing Tasks through Intelligent Agent},
  author={Xu, Wenjia and Yu, Zijian and Mu, Boyang and Wei, Zhiwei and Zhang, Yuanben and Li, Guangzuo and Peng, Mugen},
  journal={arXiv preprint arXiv:2406.07089},
  year={2024}
}

@inproceedings{zhang2024aflow,
  title={{AF}low: Automating agentic workflow generation},
  author={Zhang, Jiayi and Xiang, Jinyu and Yu, Zhaoyang and Teng, Fengwei and Chen, Xionghui and Chen, Jiaqi and Zhuge, Mingchen and Cheng, Xin and Hong, Sirui and Wang, Jinlin and others},
  booktitle={The Thirteenth International Conference on Learning Representations},
  year={2025}
}

@article{zhang2024geogpt,
  title={{GeoGPT}: An assistant for understanding and processing geospatial tasks},
  author={Zhang, Yifan and Wei, Cheng and He, Zhengting and Yu, Wenhao},
  journal={International Journal of Applied Earth Observation and Geoinformation},
  volume={131},
  pages={103976},
  year={2024},
  publisher={Elsevier}
}

@inproceedings{zhu2025knowagent,
  title={{KnowAgent}: Knowledge-augmented planning for {LLM}-based agents},
  author={Zhu, Yuqi and Qiao, Shuofei and Ou, Yixin and Deng, Shumin and Lyu, Shiwei and Shen, Yue and Liang, Lei and Gu, Jinjie and Chen, Huajun and Zhang, Ningyu},
  booktitle={Findings of the Association for Computational Linguistics: NAACL 2025},
  pages={3709--3732},
  year={2025}
}

@article{akinboyewa2025gis,
  title={{GIS} {C}opilot: Towards an autonomous GIS agent for spatial analysis},
  author={Akinboyewa, Temitope and Li, Zhenlong and Ning, Huan and Lessani, M Naser},
  journal={International Journal of Digital Earth},
  volume={18},
  number={1},
  pages={2497489},
  year={2025},
  publisher={Taylor \& Francis}
}

@article{zhang2022progress,
  title={Progress and challenges in intelligent remote sensing satellite systems},
  author={Zhang, Bing and Wu, Yuanfeng and Zhao, Boya and Chanussot, Jocelyn and Hong, Danfeng and Yao, Jing and Gao, Lianru},
  journal={IEEE Journal of Selected Topics in Applied Earth Observations and Remote Sensing},
  volume={15},
  pages={1814--1822},
  year={2022},
  publisher={IEEE}
}

@article{hu2025ringmo,
  title={Ring{M}o-{A}gent: A Unified Remote Sensing Foundation Model for Multi-Platform and Multi-Modal Reasoning},
  author={Hu, Huiyang and Wang, Peijin and Feng, Yingchao and Wei, Kaiwen and Yin, Wenxin and Diao, Wenhui and Wang, Mengyu and Bi, Hanbo and Kang, Kaiyue and Ling, Tong and others},
  journal={arXiv preprint arXiv:2507.20776},
  year={2025}
}

@article{hu2025rsgpt,
  title={{RSGPT}: A remote sensing vision language model and benchmark},
  author={Hu, Yuan and Yuan, Jianlong and Wen, Congcong and Lu, Xiaonan and Liu, Yu and Li, Xiang},
  journal={ISPRS Journal of Photogrammetry and Remote Sensing},
  volume={224},
  pages={272--286},
  year={2025},
  publisher={Elsevier}
}

@article{shabbir2025thinkgeo,
  title={Think{G}eo: Evaluating Tool-Augmented Agents for Remote Sensing Tasks},
  author={Shabbir, Akashah and Munir, Muhammad Akhtar and Dudhane, Akshay and Sheikh, Muhammad Umer and Khan, Muhammad Haris and Fraccaro, Paolo and Moreno, Juan Bernabe and Khan, Fahad Shahbaz and Khan, Salman},
  journal={arXiv preprint arXiv:2505.23752},
  year={2025}
}

@article{zhan2025skyeyegpt,
  title={Sky{E}ye{GPT}: Unifying remote sensing vision-language tasks via instruction tuning with large language model},
  author={Zhan, Yang and Xiong, Zhitong and Yuan, Yuan},
  journal={ISPRS Journal of Photogrammetry and Remote Sensing},
  volume={221},
  pages={64--77},
  year={2025},
  publisher={Elsevier}
}

@article{zhang2019remotely,
  title={Remotely sensed big data: Evolution in model development for information extraction [point of view]},
  author={Zhang, Bing and Chen, Zhengchao and Peng, Dailiang and Benediktsson, Jon Atli and Liu, Bo and Zou, Lei and Li, Jun and Plaza, Antonio},
  journal={Proceedings of the IEEE},
  volume={107},
  pages={2294--2301},
  year={2019},
  publisher={IEEE}
}

@article{zhang2025core,
  title={The core concepts and fundamental issues of remote sensing science},
  author={Zhang, Bing and Liu, Qinhuo and Li, Xiaoming and Liu, Liangyun and Yang, Bisheng and Husi, Letu and Gao, Lianru and Zhang, Wenjuan and Zhang, Hao and Bian, Zunjian and Qi, Mengjia and Chen, Chi and Shang, Huazhe},
  journal={National Remote Sensing Bulletin},
  volume={29},
  pages={1--48},
  year={2025}
}

@article{Wang2024GTA,
  author  = {Wang, Jize and Ma, Zerun and Li, Yining and Zhang, Songyang and Chen, Cailian and Chen, Kai and Le, Xinyi},
  title   = {{GTA}: a benchmark for general tool agents},
  journal = {Advances in Neural Information Processing Systems},
  year    = {2024},
  volume  = {37},
  pages   = {75749--75790}
}

@article{jiang2023mistral,
  title={Mistral 7B},
  author={Jiang, Albert Q and Sablayrolles, Alexandre and Mensch, Arthur and Bamford, Chris and Chaplot, Devendra Singh and de las Casas, Diego and Bressand, Florian and Lengyel, Gianna and Lample, Guillaume and Saulnier, Lucile and others},
  journal={arXiv preprint arXiv:2310.06825},
  year={2023}
}

@article{dubey2024llama,
  title={The {L}lama 3 herd of models},
  author={Dubey, Abhimanyu and Jauhri, Abhinav and Pandey, Abhinav and Kadian, Abhishek and Al-Dahle, Ahmad and Letman, Aiesha and Mathur, Akhil and Schelten, Alan and Yang, Amy and Fan, Angela and others},
  journal={arXiv preprint arXiv:2407.21783},
  year={2024},
  url={https://arxiv.org/abs/2407.21783},
  doi={10.48550/arXiv.2407.21783}
}

@article{guo2025deepseek,
  title={{DeepSeek-R1} incentivizes reasoning in {LLM}s through reinforcement learning},
  author={Guo, Daya and Yang, Dejian and Zhang, Haowei and Song, Junxiao and Wang, Peiyi and Zhu, Qihao and Xu, Runxin and Zhang, Ruoyu and Ma, Shirong and Bi, Xiao and others},
  journal={Nature},
  volume={645},
  pages={633--638},
  year={2025},
  publisher={Nature Publishing Group UK London}
}

@article{liu2024deepseek,
  title={{DeepSeek-V3} technical report},
  author={Liu, Aixin and Feng, Bei and Xue, Bing and Wang, Bingxuan and Wu, Bochao and Lu, Chengda and Zhao, Chenggang and Deng, Chengqi and Zhang, Chenyu and Ruan, Chong and others},
  journal={arXiv preprint arXiv:2412.19437},
  year={2024}
}

@article{achiam2023gpt,
  title={{GPT}-4 technical report},
  author={Achiam, Josh and Adler, Steven and Agarwal, Sandhini and Ahmad, Lama and Akkaya, Ilge and Aleman, Florencia Leoni and Almeida, Diogo and Altenschmidt, Janko and Altman, Sam and Anadkat, Shyamal and others},
  journal={arXiv preprint arXiv:2303.08774},
  year={2023}
}

@article{chen2025cangling,
  title={{C}ang{L}ing-{K}now{F}low: A Unified Knowledge-and-Flow-fused Agent for Comprehensive Remote Sensing Applications},
  author={Chen, Zhengchao and Wang, Haoran and Yao, Jing and Ghamisi, Pedram and Zhou, Jun and Atkinson, Peter M and Zhang, Bing},
  journal={arXiv preprint arXiv:2512.15231},
  year={2025}
}

@inproceedings{liu_toolace_2025,
title={Tool{ACE}: Winning the Points of {LLM} Function Calling},
author={Weiwen Liu and Xu Huang and Xingshan Zeng and xinlong hao and Shuai Yu and Dexun Li and Shuai Wang and Weinan Gan and Zhengying Liu and Yuanqing Yu and Zezhong WANG and Yuxian Wang and Wu Ning and Yutai Hou and Bin Wang and Chuhan Wu and Wang Xinzhi and Yong Liu and Yasheng Wang and Duyu Tang and Dandan Tu and Lifeng Shang and Xin Jiang and Ruiming Tang and Defu Lian and Qun Liu and Enhong Chen},
booktitle={The Thirteenth International Conference on Learning Representations},
year={2025},
url={https://openreview.net/forum?id=8EB8k6DdCU}
}

@article{xu_toucan_2025,
  title = {{TOUCAN}: {Synthesizing} 1.{5M} {Tool}-{Agentic} {Data} from {Real}-{World} {MCP} {Environments}},
  author = {Xu, Zhangchen and Soria, Adriana Meza and Tan, Shawn and Roy, Anurag and Agrawal, Ashish Sunil and Poovendran, Radha and Panda, Rameswar},
  year = {2025},
  journal={arXiv preprint arXiv:2510.01179},
}

@article{liu2024apigen,
  title={{APIGen}: Automated pipeline for generating verifiable and diverse function-calling datasets},
  author={Liu, Zuxin and Hoang, Thai and Zhang, Jianguo and Zhu, Ming and Lan, Tian and Tan, Juntao and Yao, Weiran and Liu, Zhiwei and Feng, Yihao and RN, Rithesh and others},
  journal={Advances in Neural Information Processing Systems},
  volume={37},
  pages={54463--54482},
  year={2024}
}

@inproceedings{wang2025instructrag,
  title={{InstructRAG}: Leveraging retrieval-augmented generation on instruction graphs for llm-based task planning},
  author={Wang, Zheng and Teo, Shu Xian and Chew, Jun Jie and Shi, Wei},
  booktitle={Proceedings of the 48th International ACM SIGIR Conference on Research and Development in Information Retrieval},
  pages={1413--1422},
  year={2025}
}

@inproceedings{lin2023ra,
title={{RA}-{DIT}: Retrieval-Augmented Dual Instruction Tuning},
author={Xi Victoria Lin and Xilun Chen and Mingda Chen and Weijia Shi and Maria Lomeli and Richard James and Pedro Rodriguez and Jacob Kahn and Gergely Szilvasy and Mike Lewis and Luke Zettlemoyer and Wen-tau Yih},
booktitle={The Twelfth International Conference on Learning Representations},
year={2024},
url={https://openreview.net/forum?id=22OTbutug9}
}

@article{kagaya2024rap,
  title={{RAP}: Retrieval-augmented planning with contextual memory for multimodal llm agents},
  author={Kagaya, Tomoyuki and Yuan, Thong Jing and Lou, Yuxuan and Karlekar, Jayashree and Pranata, Sugiri and Kinose, Akira and Oguri, Koki and Wick, Felix and You, Yang},
  journal={arXiv preprint arXiv:2402.03610},
  year={2024}
}

@inproceedings{zhang2024raft,
title={{RAFT}: Adapting Language Model to Domain Specific {RAG}},
author={Tianjun Zhang and Shishir G Patil and Naman Jain and Sheng Shen and Matei Zaharia and Ion Stoica and Joseph E. Gonzalez},
booktitle={First Conference on Language Modeling},
year={2024},
url={https://openreview.net/forum?id=rzQGHXNReU}
}

@inproceedings{jiao2025hirag,
    title = "{HIRAG}: Hierarchical-Thought Instruction-Tuning Retrieval-Augmented Generation",
    author = "Jiao, Yihan  and
      Tan, Zhehao  and
      Yang, Dan  and
      Sun, Duolin  and
      Feng, Jie  and
      Shen, Yue  and
      Wang, Jian  and
      Wei, Peng",
    editor = "Christodoulopoulos, Christos  and
      Chakraborty, Tanmoy  and
      Rose, Carolyn  and
      Peng, Violet",
    booktitle = "Findings of the Association for Computational Linguistics: EMNLP 2025",
    month = nov,
    year = "2025",
    address = "Suzhou, China",
    publisher = "Association for Computational Linguistics",
    url = "https://aclanthology.org/2025.findings-emnlp.274/",
    doi = "10.18653/v1/2025.findings-emnlp.274",
    pages = "5111--5130",
    ISBN = "979-8-89176-335-7",
}

@misc{michelakis2025core,
  title={{CORE}: {Full}-{Path} {Evaluation} of {LLM} {Agents} {Beyond} {Final} {State}},
  author={Michelakis, Panagiotis and Hadjiyiannis, Yiannis and Stamoulis, Dimitrios},
  url={http://arxiv.org/abs/2509.20998},
  doi={10.48550/arXiv.2509.20998},
  publisher={arXiv},
  year={2025},
  note={arXiv:2509.20998 [cs]}
}

@misc{feng2025earthagent,
  title={Earth-{Agent}: {Unlocking} the {Full} {Landscape} of {Earth} {Observation} with {Agents}},
  author={Feng, Peilin and Lv, Zhutao and Ye, Junyan and Wang, Xiaolei and Huo, Xinjie and Yu, Jinhua and Xu, Wanghan and Zhang, Wenlong and Bai, Lei and He, Conghui and Li, Weijia},
  url={http://arxiv.org/abs/2509.23141},
  doi={10.48550/arXiv.2509.23141},
  publisher={arXiv},
  year={2025},
  note={arXiv:2509.23141 [cs]}
}

@misc{li2025designing,
  title={Designing {Domain}-{Specific} {Agents} via {Hierarchical} {Task} {Abstraction} {Mechanism}},
  author={Li, Kaiyu and Wang, Jiayu and Wang, Zhi and Qiao, Hui and Zhang, Weizhan and Meng, Deyu and Cao, Xiangyong},
  url={http://arxiv.org/abs/2511.17198},
  doi={10.48550/arXiv.2511.17198},
  year={2025}
}

@inproceedings{krechetova2025geobenchx,
  title={{GeoBenchX}: {Benchmarking} {LLMs} in {Agent} {Solving} {Multistep} {Geospatial} {Tasks}},
  author={Krechetova, Varvara and Kochedykov, Denis},
  year={2025},
  booktitle={Proceedings of the 1st ACM SIGSPATIAL International Workshop on Generative and Agentic AI for Multi-Modality Space-Time Intelligence},
  pages={27--35},
}

@article{yang2024qwen2,
  title={Qwen2.5 Technical Report},
  author={Yang, An and Yang, Baosong and Zhang, Beichen and others},
  journal={arXiv preprint arXiv:2412.15115},
  year={2024}
}

@inproceedings{hu2022lora,
  title={{LoRA}: Low-Rank Adaptation of Large Language Models},
  author={Hu, Edward J. and Shen, Yelong and Wallis, Phillip and Allen-Zhu, Zeyuan and Li, Yuanzhi and Wang, Shean and Wang, Lu and Chen, Weizhu},
  booktitle={International Conference on Learning Representations},
  year={2022}
}

@inproceedings{patil2024gorilla,
  title={{Gorilla}: Large Language Model Connected with Massive {APIs}},
  author={Patil, Shishir G. and Zhang, Tianjun and Wang, Xin and Gonzalez, Joseph E.},
  booktitle={Advances in Neural Information Processing Systems},
  year={2024}
}

@article{qwen3,
  title={{Qwen3} Technical Report},
  author={Yang, An and Li, Anfeng and Yang, Baosong and others},
  journal={arXiv preprint arXiv:2505.09388},
  year={2025}
}

\clearpage
\appendix

\newcommand{\simcrafttoolentry}[4]{%
\par\smallskip\noindent
\begin{minipage}{\linewidth}
\raggedright\footnotesize\sloppy
\textbf{\texttt{#1}}\\[-0.15em]
\emph{Req.} #2\\[-0.10em]
\emph{Opt.} #3\\[-0.10em]
\emph{I/O.} #4
\end{minipage}
\par\smallskip
}

\section{Dataset Construction and Splits}
\label{app:dataset}

This section records the SimRS-14k construction quantities and split rules used in Section~\ref{sec:exp-setup}. SimRS-14k contains Remote Sensing (RS) tool-use trajectories generated from expert-seeded tasks. The content supports reproducibility and does not add experimental claims beyond the main paper.

\smallskip
\noindent\textbf{SimRS-14k Construction Summary.}
\begin{center}
\small
\begin{tabular*}{\columnwidth}{@{\extracolsep{\fill}}p{2.55cm}rp{2.05cm}@{}}
\toprule
\textbf{Item} & \textbf{Value} & \textbf{Role} \\
\midrule
Expert-seeded tasks & 162 & Initial task pool \\
Seed-task application domains & 9 & Intent coverage \\
Seed-task technical categories & 8 & Operation coverage \\
Candidate trajectories & 16,200 & Before validation \\
Rejected candidates & 2197 & Failed validation \\
Retained trajectories & 14,003 & SimRS-14k corpus \\
Expert-mode trajectories & 6987 & Complete initial requests \\
Novice-mode trajectories & 7016 & Clarification required \\
CRAFT training trajectories & 13,503 & Fine-tuning split \\
Held-out test trajectories & 500 & SimRS-14k test split \\
\bottomrule
\end{tabular*}
\end{center}

The nine application domains and eight technical categories describe seed-task coverage, not the five functional domains used to organize the Atomic Toolset. We sample the test split with stratification over seed-task type, application domain, and trajectory length. The held-out split shares no geographic coordinates, time windows, or sensor parameters with the training split. This design reduces within-synthesis leakage and tests whether the model learns reusable tool-composition patterns rather than repeated locations, dates, or sensor settings. The novice-ratio ablation in the main paper uses a fixed 13,974-trajectory subset, equal to twice the smaller post-filtered mode count ($2 \times 6987$), so expert-only, novice-only, and mixed-ratio settings are compared at the same corpus size.

\section{Atomic Toolset Schemas and I/O Specifications}
\label{app:toolset}

This section provides the full Atomic Toolset specification referenced in Section~\ref{sec:toolset}. The 36 RS-specific tools are mapped exhaustively to the five functional domains used in the main paper. Each schema entry is rendered in compact prose: required arguments are mandatory JSON keys, optional arguments are typed optional keys, and the I/O field records the accepted artifact type and registered output returned to the Dynamic Path Registry.

The schema checker validates required-key presence, JSON type, enumerated values, numeric ranges when specified, and artifact compatibility. Path-valued arguments must refer to artifacts returned by previous valid tool calls as \texttt{downloaded\_files}, \texttt{output\_path}, or \texttt{output\_paths}. Raster-consuming tools accept GeoTIFF-like raster artifacts, vector-consuming tools accept GeoJSON or Shapefile-like vector artifacts, and visualization tools accept registered raster, vector, metadata, or statistics artifacts. The inventory below uses \emph{Req.}, \emph{Opt.}, and \emph{I/O} to denote required arguments, optional arguments, and registered input-output artifacts.

\subsection{Domain Mapping}

The five-domain mapping is exhaustive: Data \& Preprocessing contains 12 tools, AI Interpretation contains 6 tools, Spatio-Temporal Analysis contains 4 tools, Physical \& GIS Analytics contains 11 tools, and Visualization contains 3 tools. This gives the 36-tool action space $\mathcal{A}$ used in the main paper.

\smallskip
\noindent\textbf{Representative Tool Summary.}
\begin{center}
\scriptsize
\setlength{\tabcolsep}{3pt}
\renewcommand{\arraystretch}{1.08}
\begin{tabular}{@{}p{1.65cm}p{5.15cm}@{}}
\toprule
\textbf{Domain} & \textbf{Representative Tools} \\
\midrule
Data \& Preprocessing & \texttt{rs\_data\_search}, \texttt{rs\_data\_download}, \texttt{rs\_geo\_reproject}, \texttt{rs\_pro\_radiometric\_calibration}. \\
AI Interpretation & \texttt{rs\_ai\_object\_detection}, \texttt{rs\_ai\_semantic\_segmentation}, \texttt{rs\_ai\_change\_detection}. \\
Spatio-Temporal Analysis & \texttt{rs\_temporal\_statistics}, \texttt{rs\_temporal\_trend\_analysis}, \texttt{rs\_temporal\_disturbance\_detection}. \\
Physical \& GIS Analytics & \texttt{rs\_calc\_spectral\_index}, \texttt{rs\_gis\_overlay}, \texttt{rs\_sar\_processing}, \texttt{rs\_lidar\_processing}. \\
Visualization & \texttt{rs\_vis\_render\_map}, \texttt{rs\_vis\_plot\_chart}, \texttt{rs\_raster\_to\_vector}. \\
\bottomrule
\end{tabular}
\end{center}

\subsection{Data and Preprocessing Schemas}

\simcrafttoolentry{rs\_data\_search}
{\texttt{location:string}; \texttt{time\_range:array[string]}.}
{\texttt{time\_step:enum} with year, quarter, month, or first\_last; \texttt{platform:string}; \texttt{max\_cloud\_cover:integer} in $[0,100]$.}
{Inputs are query slots for place, time, platform, and cloud cover; output is a metadata record set with searchable \texttt{image\_ids}.}

\simcrafttoolentry{rs\_data\_download}
{\texttt{image\_ids:array[string]}.}
{\texttt{output\_dir:string}.}
{Input is the image ID list returned by search; output is \texttt{downloaded\_files}, which registers raster artifacts for downstream tools.}

\simcrafttoolentry{rs\_geo\_reproject}
{\texttt{image\_paths:array[string]}; \texttt{target\_crs:string}.}
{None.}
{Inputs are registered raster paths; output is reprojected raster artifact paths.}

\simcrafttoolentry{rs\_geo\_crop}
{\texttt{image\_paths:array[string]}.}
{\texttt{roi\_bbox:array}; \texttt{vector\_mask\_path:string}.}
{Inputs are registered rasters and an optional vector boundary; output is cropped raster artifact paths.}

\simcrafttoolentry{rs\_geo\_mosaic}
{\texttt{image\_paths:array[string]}.}
{\texttt{overlap\_method:enum} with mean, first, or blend.}
{Inputs are adjacent registered rasters; output is a mosaicked raster artifact path.}

\simcrafttoolentry{rs\_geo\_registration}
{\texttt{src\_image\_path:string}; \texttt{ref\_image\_path:string}.}
{None.}
{Inputs are source and reference raster artifacts; output is a registered raster artifact path.}

\simcrafttoolentry{rs\_pro\_radiometric\_calibration}
{\texttt{image\_paths:array[string]}; \texttt{calibration\_level:enum} with TOA\_Reflectance, Surface\_Reflectance, or Brightness\_Temp.}
{\texttt{sensor:string}.}
{Inputs are registered rasters and optional sensor metadata; output is calibrated raster artifact paths.}

\simcrafttoolentry{rs\_atmos\_cloud\_remove}
{\texttt{image\_paths:array[string]}.}
{\texttt{method:string} with dark\_channel or qa\_band\_mask.}
{Inputs are registered raster paths; output is cloud- or haze-corrected raster artifact paths.}

\simcrafttoolentry{rs\_radio\_enhance}
{\texttt{image\_paths:array[string]}; \texttt{operation:string} with stretch, equalize, or pansharpen.}
{\texttt{pan\_path:string}.}
{Inputs are registered rasters and an optional panchromatic image; output is enhanced raster artifact paths.}

\simcrafttoolentry{rs\_utils\_read\_metadata}
{\texttt{input\_paths:array[string]}.}
{None.}
{Inputs are registered image paths; output is metadata covering bands, resolution, CRS, and spatial extent.}

\simcrafttoolentry{rs\_utils\_file\_convert}
{\texttt{input\_paths:array[string]}; \texttt{target\_format:string}.}
{None.}
{Inputs are registered raster or vector files; output is converted artifact paths in the requested format.}

\simcrafttoolentry{rs\_utils\_coord\_transform}
{\texttt{input:string}.}
{\texttt{ref\_image:string}.}
{Input is a place name or pixel coordinate, with an optional reference image; output is a coordinate object usable by later spatial filters.}

\subsection{AI Interpretation Schemas}

\simcrafttoolentry{rs\_ai\_object\_detection}
{\texttt{image\_paths:array[string]}; \texttt{target\_class:string}.}
{\texttt{confidence\_threshold:number} in $[0,1]$.}
{Inputs are registered rasters; output is a detection artifact with target locations and confidence scores.}

\simcrafttoolentry{rs\_ai\_semantic\_segmentation}
{\texttt{image\_paths:array[string]}; \texttt{target\_class:string}.}
{\texttt{target\_classes:array[string]}.}
{Inputs are registered rasters; output is a class-specific mask artifact.}

\simcrafttoolentry{rs\_ai\_land\_use\_classification}
{\texttt{image\_paths:array[string]}.}
{\texttt{scheme:string}.}
{Inputs are registered rasters; output is a full-coverage classification raster and a \texttt{class\_map} for later visualization or vectorization.}

\simcrafttoolentry{rs\_ai\_instance\_segmentation}
{\texttt{image\_paths:array[string]}; \texttt{target\_class:string}.}
{None.}
{Inputs are registered rasters; output is an instance mask or contour artifact for counting and geometry analysis.}

\simcrafttoolentry{rs\_ai\_change\_detection}
{\texttt{image\_path\_t1:string}; \texttt{image\_path\_t2:string}.}
{\texttt{mode:enum} with binary or semantic.}
{Inputs are registered pre-change and post-change rasters; output is a change mask or semantic change artifact.}

\simcrafttoolentry{rs\_ai\_scene\_classification}
{\texttt{image\_paths:array[string]}.}
{None.}
{Inputs are registered rasters; output is a scene-label result object and optional summary artifact.}

\subsection{Spatio-Temporal Analysis Schemas}

\simcrafttoolentry{rs\_temporal\_statistics}
{\texttt{image\_paths:array[string]}; \texttt{statistic\_type:string}.}
{\texttt{image\_stack\_dir:string}; \texttt{target\_band:string}.}
{Inputs are time-series raster artifacts; output is a statistics object or file with the requested aggregation.}

\simcrafttoolentry{rs\_temporal\_preprocessing}
{\texttt{image\_paths:array[string]}; \texttt{operation:string} with smooth or interpolate.}
{\texttt{image\_stack\_dir:string}; \texttt{params:object}.}
{Inputs are time-series raster artifacts; output is a smoothed or gap-filled raster stack.}

\simcrafttoolentry{rs\_temporal\_trend\_analysis}
{\texttt{image\_paths:array[string]}; \texttt{method:string} with linear\_regression or mann\_kendall.}
{\texttt{image\_stack\_dir:string}; \texttt{output\_metrics:array[string]} such as slope or p\_value.}
{Inputs are time-series raster artifacts; output is a trend result artifact with requested metrics.}

\simcrafttoolentry{rs\_temporal\_disturbance\_detection}
{\texttt{image\_paths:array[string]}; \texttt{algorithm:string} with bfast or landtrendr.}
{\texttt{image\_stack\_dir:string}; \texttt{params:object}.}
{Inputs are time-series raster artifacts; output is a disturbance or breakpoint artifact.}

\subsection{Physical and GIS Analytics Schemas}

\simcrafttoolentry{rs\_calc\_spectral\_index}
{\texttt{image\_paths:array[string]}; \texttt{index\_name:string}.}
{None.}
{Inputs are registered rasters with required spectral bands; output is an index raster artifact.}

\simcrafttoolentry{rs\_calc\_parameter\_inversion}
{\texttt{image\_paths:array[string]}; \texttt{parameter:string}.}
{None.}
{Inputs are registered rasters with parameter-compatible bands or products; output is a physical-parameter raster artifact.}

\simcrafttoolentry{rs\_raster\_math\_calc}
{\texttt{raster\_paths:array[string]}; \texttt{mode:string} with threshold\_mask or band\_math; \texttt{expression:string}.}
{\texttt{output\_type:enum} with mask or value.}
{Inputs are registered rasters; output is a derived raster, mask, or scalar result artifact.}

\simcrafttoolentry{rs\_surface\_geometry\_analysis}
{\texttt{scalar\_field\_paths:array[string]}; \texttt{operation:string} with slope, aspect, or hillshade.}
{\texttt{z\_factor:number}; \texttt{params:object}.}
{Inputs are DEM-like scalar fields; output is a terrain-analysis raster artifact.}

\simcrafttoolentry{rs\_gis\_buffer}
{\texttt{input\_vectors:array[string]}; \texttt{distance:number}.}
{\texttt{dissolve:boolean}.}
{Inputs are registered vector artifacts; output is a buffered vector artifact.}

\simcrafttoolentry{rs\_gis\_overlay}
{\texttt{input\_vectors:array[string]}; \texttt{overlay\_vector:string}; \texttt{method:enum} with intersection, union, difference, or clip.}
{None.}
{Inputs are registered vector artifacts; output is an overlay vector artifact.}

\simcrafttoolentry{rs\_gis\_zonal\_statistics}
{\texttt{image\_paths:array[string]}; \texttt{zone\_vector:string}.}
{\texttt{stats:array[string]} such as mean, max, or sum.}
{Inputs are raster artifacts and a zone vector; output is a zonal-statistics object or file.}

\simcrafttoolentry{rs\_vector\_analysis}
{\texttt{input\_vectors:array[string]}; \texttt{operation:string} with calculate\_area or count\_features.}
{None.}
{Inputs are registered vector artifacts; output is a vector-statistics object or file.}

\simcrafttoolentry{rs\_sar\_processing}
{\texttt{sar\_paths:array[string]}; \texttt{operation:string} with speckle\_filter or terrain\_correction.}
{\texttt{params:object}.}
{Inputs are SAR raster artifacts; output is a SAR-derived raster artifact.}

\simcrafttoolentry{rs\_hyperspectral\_processing}
{\texttt{image\_paths:array[string]}; \texttt{operation:string} with pca, mnf, or spectral\_unmixing.}
{\texttt{target\_components:integer}.}
{Inputs are hyperspectral raster artifacts; output is a reduced or unmixed hyperspectral artifact.}

\simcrafttoolentry{rs\_lidar\_processing}
{\texttt{point\_cloud\_paths:array[string]}; \texttt{operation:string} with point\_to\_raster or classify\_ground.}
{\texttt{resolution:number}.}
{Inputs are registered point-cloud artifacts; output is a rasterized or classified LiDAR artifact.}

\subsection{Visualization Schemas}

\simcrafttoolentry{rs\_vis\_render\_map}
{\texttt{base\_images:array[string]}.}
{\texttt{overlay\_layers:array[string]}; \texttt{class\_mapping:object}; \texttt{bands:array[integer]}; \texttt{style:string}.}
{Inputs are registered raster, vector, and class-map artifacts; output is a rendered map artifact.}

\simcrafttoolentry{rs\_vis\_plot\_chart}
{\texttt{data\_files:array[string]}; \texttt{chart\_type:string}.}
{None.}
{Inputs are registered statistics files or result objects; output is a chart artifact.}

\simcrafttoolentry{rs\_raster\_to\_vector}
{\texttt{image\_paths:array[string]}.}
{\texttt{class\_values:array[integer]}; \texttt{simplify\_tolerance:number}; \texttt{output\_format:string} with geojson or shp.}
{Inputs are registered raster masks; output is a vector artifact registered for visualization or GIS tools.}

\section{Mock Execution Engine Validation}
\label{app:mock-engine}

This section expands the planning-level checks applied by the Mock Execution Engine. The engine filters synthesized trajectories before they become supervised training targets. It does not invoke GDAL, SNAP, or Google Earth Engine on raw satellite imagery. It also does not evaluate downstream physical accuracy, such as change-detection masks or crop-classification maps.

\paragraph{Schema and Parameter Legality.} The engine rejects tool arguments that violate type, required-field, enum, or value-range constraints. Example failures include a cloud-cover threshold outside $[0,100]$, a missing \texttt{time\_range}, or a string passed to a numeric argument.

\paragraph{Dependency and Path Consistency.} The engine rejects a tool call when it references an artifact that no previous valid step produced. The Dynamic Path Registry rejects an input such as \texttt{result\_step3.tif} when no prior step registered that path.

\paragraph{Sensor and Tool Compatibility.} The engine rejects tool calls whose sensor or product is incompatible with the operation, such as a spectral index on a product lacking a near-infrared band.

\paragraph{Error Injection.} Beyond the three validation checks, the engine injects recoverable runtime-like failures to test replanning behavior. For example, it can emit \texttt{CloudCoverExceeded} and route the failure to the Reflection Agent for an adjusted search or preprocessing plan.

The engine maintains a Dynamic Path Registry $\mathcal{R}$ during synthesis. When a simulated tool call succeeds, the engine registers its output artifacts and exposes them to subsequent steps. When a later tool call references an unregistered artifact, the engine returns \texttt{PathNotFoundError}. This mechanism prevents trajectories from containing fabricated intermediate files that would otherwise appear syntactically plausible.

The error-injection check uses probability $\varepsilon_{\text{exec}}{=}0.15$ in Phase~I. The main paper sets the Phase~I error-injection probability and the Phase~II context-perturbation probability to $\varepsilon{=}0.15$. This appendix uses subscripted notation only to distinguish the two roles. Phase~I injects execution-like failures during data synthesis, whereas Phase~II perturbs retrieved SOP context during fine-tuning.

\smallskip
\noindent\textbf{Candidate Filtering Summary.}
\begin{center}
\small
\begin{tabular*}{\columnwidth}{@{\extracolsep{\fill}}lrr@{}}
\toprule
\textbf{Status} & \textbf{Count} & \textbf{Share} \\
\midrule
Candidate trajectories & 16,200 & 100.0\% \\
Rejected candidates & 2197 & 13.6\% \\
Retained trajectories & 14,003 & 86.4\% \\
\bottomrule
\end{tabular*}
\end{center}

\section{Multi-Agent Synthesis Details}

\begin{table}[t!]
\centering
\small
\renewcommand{\arraystretch}{1.15}
\begin{tabular*}{\columnwidth}{@{\extracolsep{\fill}}lcc@{}}
\toprule
\textbf{Metric} & \textbf{Estimate (\%)} & \textbf{95\% CI} \\
\midrule
PSR & 83.2 & [80.6, 85.7] \\
TSA & 93.5 & [91.8, 95.0] \\
AEM & 79.6 & [76.8, 82.3] \\
\bottomrule
\end{tabular*}
\caption{\textbf{Bootstrap confidence intervals for SimCRAFT-Qwen2.5-7B} on the SimRS-14k test split (1000 resamples).}
\label{tab:bootstrap-ci}
\end{table}

\label{app:synthesis}

Phase~I combines user-intent diversification with role-specialized trajectory generation. This section separates user simulation from trajectory synthesis, because dialogue clarification and tool-chain generation have different activation conditions.

\paragraph{User Simulation Agent.} Before trajectory synthesis, the User Simulation Agent produces the initial user request. Expert Mode states all required slots, whereas Novice Mode masks a random subset of critical slots.

\paragraph{Clarification Agent.} When a critical slot is missing, the Clarification Agent maintains a slot ledger for time range, sensor, and spatial extent. It asks targeted follow-up questions until the clarified intent $q^*$ is complete.

\paragraph{Planning Agent.} In the initial synthesis state, the Planning Agent drafts a high-level RS workflow roadmap constrained by the Atomic Toolset.

\paragraph{Execution Agent.} In the normal execution state, the Execution Agent emits the next atomic tool call $(t_i,\boldsymbol{\theta}_i)$ from the current execution context and registered artifacts.

\paragraph{Reflection Agent.} In the failure state, the Reflection Agent receives validation or injected error messages and revises the plan before execution resumes.

\paragraph{Summary Agent.} In the completion state, the Summary Agent produces the final task summary once the trajectory reaches \texttt{Terminate}.

The shared state machine follows four stages: plan, execute, reflect, and summarize. The Planning Agent initializes the workflow, the Execution Agent proposes concrete tool calls, and the Mock Execution Engine validates each call. The Reflection Agent is dispatched only when validation fails or a recoverable error is injected. This design prevents a local invalid step from propagating through the rest of a long-horizon trajectory.

\section{CRAFT Retrieval and Fine-Tuning Details}
\label{app:craft-details}

Contextual Retrieval-Augmented Fine-Tuning (CRAFT) differs from vanilla supervised fine-tuning (SFT) by injecting retrieved Standard Operating Procedure (SOP) context from the Procedural Knowledge Base (PKB) during training. Before optimization, CRAFT perturbs the retrieved context and trains the model to recover the validated target trajectory. This section records the retrieval and optimization details reported in the main paper.

\paragraph{Retrieval Configuration.} The PKB contains 1008 SOPs. CRAFT uses a frozen dual encoder to keep train-time and inference-time retrieval behavior aligned. It filters retrieved SOPs with similarity threshold $\delta{=}0.6$ and supplies top-$k{=}2$ contexts to the model.

\paragraph{Training Objective.} The training target is the validated trajectory $y$. The model is trained to recover the correct tool sequence despite noisy retrieved context.

\paragraph{Noise-Robust Context Perturbations.} Irrelevant Context Injection teaches the model when to ignore an unrelated SOP. Parameter Mutation teaches the model to verify sensor, temporal, and argument details rather than copy them. The context perturbation probability is $\varepsilon_{\text{ctx}}{=}0.15$, which uses the same numerical value as $\varepsilon{=}0.15$ in the main paper; the subscript distinguishes Phase~II context perturbation from Phase~I error injection.

\paragraph{Optimization Setup.} The primary backbone is Qwen-2.5-7B-Instruct. Fine-tuning uses LoRA with $r{=}64$, $\alpha{=}128$, and dropout 0.05. Optimization uses AdamW with learning rate $2{\times}10^{-5}$, a cosine schedule, and 3 epochs.

At training time, the perturbation function turns the retrieved SOPs into a noisy context $\tilde{\mathcal{C}}_x$; at inference the same retriever runs without perturbation, so the model applies its learned context-denoising to clean SOPs.

\section{Bootstrap Confidence Intervals}
\label{app:ci}

We assess the stability of SimCRAFT-Qwen2.5-7B with a non-parametric bootstrap: across 1000 resamples of the test set with replacement, we report each metric's 2.5th/97.5th percentiles as a 95\% CI (Table~\ref{tab:bootstrap-ci}). The narrow intervals indicate stable estimates rather than test-split artifacts.

\end{document}